%% file: iclr2023_conference.tex
\documentclass{article} 
\usepackage{iclr2023_conference,times}

\input{math_commands.tex}

\usepackage{hyperref}
\usepackage{url}

\usepackage{graphicx}
\usepackage{booktabs}
\usepackage{subfigure}

\title{ImaginaryNet: Learning Object Detectors without Real Images and Annotations}

\author{Minheng Ni, Zitong Huang, Kailai Feng \& Wangmeng Zuo\thanks{ Corresponding Author.} \\
Faculty of Computing\\
Harbin Institute of Technology\\
\texttt{\{mhni, zthuang, klfeng\}@stu.hit.edu.cn} \quad \texttt{wmzuo@hit.edu.cn}\\
}

\iclrfinalcopy 
\begin{document}

\maketitle

\input{./Content/00_Abstract.tex}
\input{./Content/01_Introduction.tex}
\input{./Content/02_Related_Work.tex}
\input{./Content/03_Methodology.tex}
\input{./Content/04_Experiment.tex}
\input{./Content/05_Conclusion.tex}

\bibliography{iclr2023_conference}
\bibliographystyle{iclr2023_conference}

\newpage

\appendix

\input{./Content/0A_Appendix}

\end{document}

%% file: math_commands.tex
\usepackage{amsmath,amsfonts,bm}

\def\eqref#1{equation~\ref{#1}}

\def\1{\bm{1}}

\DeclareMathAlphabet{\mathsfit}{\encodingdefault}{\sfdefault}{m}{sl}
\SetMathAlphabet{\mathsfit}{bold}{\encodingdefault}{\sfdefault}{bx}{n}



%% file: Content/00_Abstract.tex
\begin{abstract}
Without the demand of training in reality, humans can easily detect a known concept simply based on its language description.
Empowering deep learning with this ability undoubtedly enables the neural network to handle complex vision tasks, \textit{e.g.}, object detection, without collecting and annotating real images.
To this end, this paper introduces a novel challenging learning paradigm Imaginary-Supervised Object Detection (ISOD), where neither real images nor manual annotations are allowed for training object detectors.
To resolve this challenge, we propose \textsc{ImaginaryNet}, a framework to synthesize images by combining pretrained language model and text-to-image synthesis model.
Given a class label, the language model is used to generate a full description of a scene with a target object, and the text-to-image model deployed to generate a photo-realistic image. 
With the synthesized images and class labels, weakly supervised object detection can then be leveraged to accomplish ISOD.
By gradually introducing real images and manual annotations, \textsc{ImaginaryNet} can collaborate with other supervision settings to further boost detection performance. 
Experiments show that \textsc{ImaginaryNet} can
(i) obtain about 70\% performance in ISOD compared with the weakly supervised counterpart of the same backbone trained on real data, (ii) significantly improve the baseline while achieving state-of-the-art or comparable performance by incorporating \textsc{ImaginaryNet} with other supervision settings.  Our code is publicly available at \href{https://github.com/kodenii/ImaginaryNet}{https://github.com/kodenii/ImaginaryNet}.
\end{abstract}

%% file: Content/01_Introduction.tex
\section{Introduction}

Humans can easily detect a known concept from language description without the demand of training in reality.
Equipping this ability to deep learning may allow the neural network to handle complex vision tasks, \textit{e.g.}, object detection, without real images and annotations.
Recently, we witness the rise of Contrastive Language-Image Pre-training (CLIP)~\citep{radford2021learning}, where general knowledge can be learned by pre-training and then be applied to various downstream tasks via zero-shot learning or task-specific fine-tuning. 
Unlike image classification, object detection is more challenging and has a larger gap than the pre-training tasks. 
%
%
Several methods, such as RegionCLIP \citep{Zhong_2022_CVPR}, ViLD \citep{gu2021open}, and Detic \citep{zhou2022detecting}, have been suggested to transfer knowledge from pre-trained CLIP \citep{radford2021learning} model to some modules of object detectors. 
However, real images and annotations are still required for some key modules of the object detectors, such as RPNs or RoI heads. 


In this work, we aim to raise and answer a question: given suitable pre-trained models, can we learn object detectors without real images and manual annotations?
To this end, we introduce a novel learning paradigm, \textit{i.e.}, Imaginary-Supervised Object Detection (ISOD), where no real images and manual annotations can be used for training object detection.
Fortunately, benefited from the progress in vision-language pre-training, ISOD is practically feasible. 
%
%
%
%
%
%
Here we propose \textsc{ImaginaryNet}, a framework to learn object detectors by combining pretrained language model as well as text-to-image
synthesis models. 
In particular, the text-to-image synthesis model is adopted to generate photo-realistic images, and the language model can be used to improve the diversity and provides class labels for the synthesized images. 
Then, ISOD can be conducted by leveraging weakly supervised object detection (WSOD) algorithms on the synthesized images with class labels to learn object detectors.
%
%
We set up a strong CLIP-based model as the baseline to verify the effectiveness of \textsc{ImaginaryNet}. Experiments show that \textsc{ImaginaryNet} can outperform the CLIP-based model with a large margin. Moreover, \textsc{ImaginaryNet} obtain about 70\% performance in ISOD compared with the weakly supervised model of the same backbone trained on real data, clearly showing the feasibility of learning object detection without any real images and manual annotations.


 
By gradually introducing real images and manual annotations, \textsc{ImaginaryNet} can collaborate with other supervision settings to further boost detection performance. 
It is worthy noting that the performance of existing object detection models may be constrained by the limited amount of training data. As a result, we use \textsc{ImaginaryNet} as a dataset expansion approach to incorporate with real images and manual annotations. 
Further experiments show that \textsc{ImaginaryNet} significantly improves the performance of the baselines while achieving state-of-the-art or comparable performance in the supervision setting. 

To sum up, the contributions of this work are as follows:
\begin{itemize}
\item We propose \textsc{ImaginaryNet}, a framework to generate synthesized images as well as supervision information for training object detector. To the best of our knowledge, we are among the first work to train deep object detectors solely based on synthesized images.
\item We propose a novel paradigm of object detection, Imaginary-Supervised Object Detection (ISOD), where no real images and annotations can be used for training object detectors. We set up the benchmark of ISOD and obtain about 70\% performance in ISOD when compared with the WSOD model of the same backbone trained on real data.
\item By incorporating with real images and manual annotations, ImaginaryNet significantly improves the baseline model while achieving state-of-the-art or comparable performance.
\end{itemize}



%% file: Content/02_Related_Work.tex
\section{Related Work}

\subsection{Object Detection}

Most fully-supervised object detection methods (FSOD) \citep{ren2015faster,redmon2016you,tian2019fcos,carion2020end} rely on large amount of training data with box-level annotations. To reduce the annotation costs, some works attempt to train a detector with incompletely supervised training data. For example, weakly-supervised object detection (WSOD) \citep{huang2022w2n,dong2021boosting,bilen2016weakly,tang2017multiple} requires only image-level labels. While semi-supervised object detection (SSOD) \citep{liu2021unbiased,xu2021end,chen2022label} leverages unlabeled data combining with box-level labeled data. Although these works use relatively less or weak supervision, all of these works still rely on real images and manual annotations.  In this paper, we propose ISOD, where no real images and manual annotations can be used for training object detection, thereby saving the demand for data acquisition and annotation costs.

\subsection{Sim2Real}

Many works explore to use of simulation to learn real skills for models, especially in robotics. Some works \citep{horvath2022object, akhyani2022towards, sadeghi2016cad2rl} also explore the feasibility of combining simulation in computer vision tasks. \cite{wang2018low} use simulated images to train the classification model. Recently, \cite{ge2022dall} used a simulated scene with real foreground object images to learn object detection. However, due to the domain gap between simulated and real images, these models still need real images or annotations to transfer knowledge to the real domain. 

\subsection{Pre-trained Models}

Pre-trained models have shown effectiveness in many vision tasks including object detection.
Some CLIP-based models can transfer knowledge from pre-trained CLIP \citep{radford2021learning} model. However, real images and manual annotations are still needed for their RPNs or RoI heads.
Visual synthesis models, such as GAN \citep{goodfellow2016deep}, StyleGAN \citep{karras2019style}, and ImageBART \cite{esser2021imagebart} aim to generate plausible images.
In recent years, we witness the rise of text-to-image synthesis models with high quality and capability of language controlling, such as DALL-E \citep{ramesh2021zero}, Stable Diffusion \citep{rombach2022high}, and Imagen \citep{saharia2022photorealistic}.
However, little studies have been given to only use the knowledge from pre-trained text-to-image synthesis models to handle complex vision tasks, \textit{e.g.}, object detection.

%% file: Content/03_Methodology.tex
\section{Methodology}

\subsection{Preliminaries and Problem Formulations}

Object detection aims to find each object with the bounding box $\mathbf{b}_k \in \mathbb{R}^4$ and  class label $c_k \in \mathbb{C}$ in a given image $\mathbf{I} \in \mathbb{R}^{C\times W\times H}$, where $C$, $W$ and $H$ denote the channels, width and height of the image and $k$ denotes the $k$-th object, respectively. In standard training process of object detection, $\mathcal{D}^{\mathrm{R}}$ is the dataset which contains real images and their corresponding annotations.
In practice, $\mathcal{D}^{\mathrm{R}}$ may contain different types of data and annotations, where $\mathcal{D}^{\mathrm{R}} = \mathcal{D}^{\mathrm{R}}_s \cup\mathcal{D}^{\mathrm{R}}_w\cup\mathcal{D}^{\mathrm{R}}_u$.
Class-labeled real data $\mathcal{D}^{\mathrm{R}}_w = \{(\mathbf{I}, \{(c)_i\})_k\}$ contains real images and corresponding class labels, which is used for WSOD setting.
Box-labeled real data $\mathcal{D}^{\mathrm{R}}_s = \{(\mathbf{I}, \{(\mathbf{b}, c)_i\})_k\}$ contains real images, bounding box and class label of each object.
Un-annotated real data $\mathcal{D}^{\mathrm{R}}_u = \{(\mathbf{I})_k\}$ only contains real images. Box-level annotated and un-annotated real data are used in SSOD setting. 


In this paper, we propose Imaginary-Supervised Object Detection (ISOD), where no real images and annotations can be accessed, \textit{i.e.}, $\mathcal{D}^{\mathrm{R}} = \emptyset$.

\subsection{Imaginary Generator}

\begin{figure*}
	\centering
	\includegraphics[width=14cm]{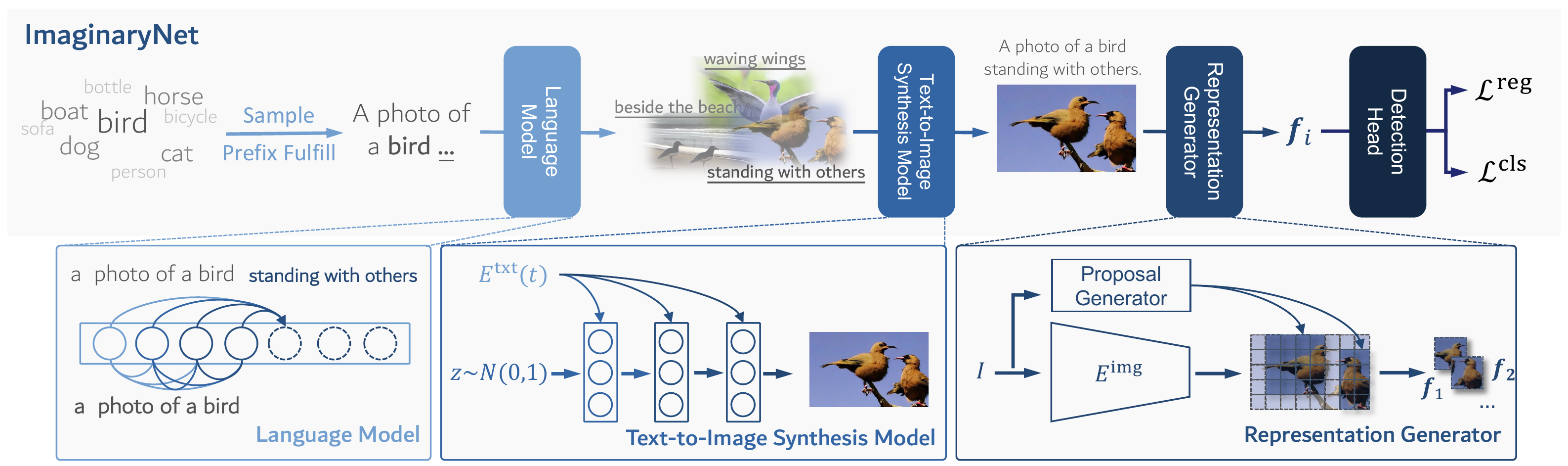}
	\caption{\textbf{Overview of \textsc{ImaginaryNet}.} \textsc{ImaginaryNet} samples the class label randomly and fulfills it to the prefix template. The language model extends the prefix to a complete description. The synthesis model generates imaginary images from random noise based on the description. Proposal representations are extracted from imaginary images. \textsc{ImaginaryNet} optimizes Detection Head with proposal representations and class labels. If real data exists, Detection Head will also be optimized based on representations from real images and manual annotations.}
	\label{fig:model}
\end{figure*}

\textsc{ImaginaryNet} aims to generate synthesized images as well as supervision information for training object detectors.
Given a class vocab $\mathbb{C}$, \textsc{ImaginaryNet} samples uniformly to get a class label $c\in \mathbb{C}$. Then, the class name of $c$ will be fulfilled into a prefix phrase, which is "\textit{A photo of a}", to obtain the subject description $P_c$ of the imaginary image.

To make the scene of image more specific and diverse, $P_c$ will be the guidance prefix of the language model $G^{\mathrm{L}}$, \textit{e.g.}, GPT-2 \citep{radford2019language}, to imagine the full description of the scene of image. 
The language model $G^{\mathrm{L}}$, which is a transformer-based autoregressive network, generates the full description token by token until meeting a full stop or exceeding the maximum of length $M^{\mathrm{L}}$ \footnote{We set $M^{\mathrm{L}}$ as 15. We tried different $M^{\mathrm{L}}$, but no significant gain was obtained.}.

Let $\textbf{t}_c = G^{\mathrm{L}}(P_c)$ be the generated full description of the imaginary image. \textsc{ImaginaryNet} further obtain the synthesized image by 
\begin{equation}
    \textbf{I} = G^{\mathrm{V}}(z|\textbf{t}_c)
    \label{I},
\end{equation}
where $z\sim\mathcal{N}(0,1)$ is random noise sampled from normal distribution.
The text-to-image synthesis model $G^{\mathrm{V}}$ usually adopts a transformer-based autoregressive model or diffusion network, and generates  imaginary image $\textbf{I}$ based on $z$ with the guidance of $\textbf{t}_c$ in a step by step manner.

For the imaginary image $\textbf{I}$, \textsc{ImaginaryNet} obtain its feature map $\textbf{h} = E^{\mathrm{I}}(\textbf{I})$, where $E^{\mathrm{I}}$ is the image encoder consisting of multi-layers of convolutional networks. Simultaneously, proposals $\mathcal{P} = R(\textbf{I})$ of the imaginary image $\textbf{I}$ are extracted by proposal generator $R$, which adopts a  selective search module or a RPN network without initilization. \textsc{ImaginaryNet} apply RoI Pooling to each proposal $p_i \in \mathcal{P}$ in feature map $h$ to obtain the proposal representation $\textbf{f}_i \in \mathbb{R}^d$.
Consequently, \textsc{ImaginaryNet} obtain a set of imaginary proposal representations $\{\textbf{f}_i\}$ with their corresponding class-label $c$, which allows ISOD to learn object detectors without real images and annotations.

\begin{figure*}
	\centering
	\includegraphics[width=14cm]{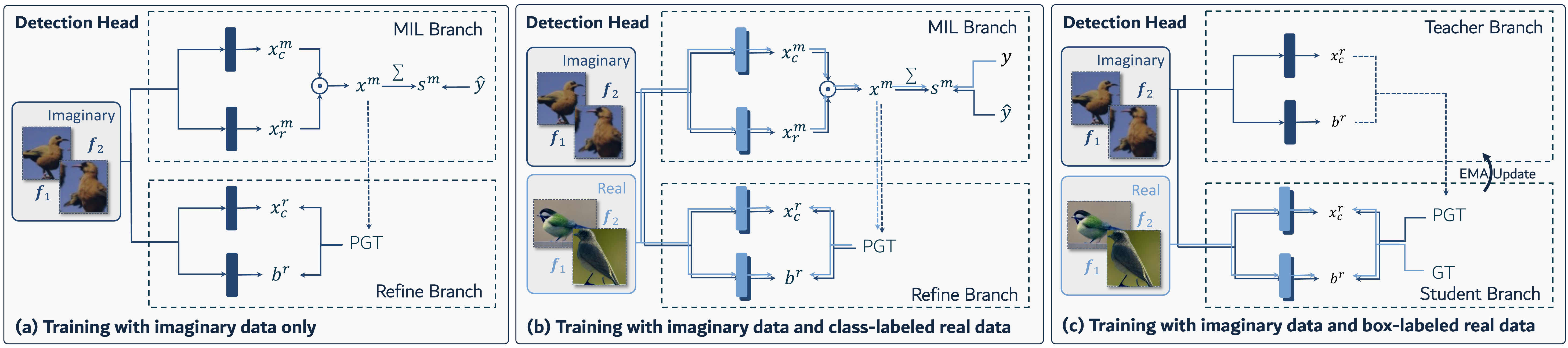}
	\caption{\textbf{The structure of Detection Head.} Based on different data settings, we use three types of detection heads. In (a) type of Detection Head, no real images and annotations will participate the training process. In types (b) and (c), real images and annotations will be trained with imaginary data together. In type (c), MIL Branch will be disabled due to box supervision exists.}
	\label{fig:roi-head}
\end{figure*}

\subsection{Detection Network}
We add an object detection head at the end of the \textbf{Representation Generator}, which takes the representation $\textbf{f}_i$ as input and predicts the bounding box of an object. 
In standard object detection, training images and corresponding annotations are given, formulating a fully-supervised training paradigm (FSOD). 
Instead, we propose a new learning paradigm named ISOD, in which neither real training images nor manual annotations can be reached. This means that we cannot apply the traditional approach \citep{ren2015faster, tian2019fcos} straightforwardly. Fortunately, our  \textsc{ImaginaryNet} samples a target class $c$  describing the semantics of the corresponding imaginary image.
Note that both the set of proposal representation $\{\textbf{f}_i\}$ and image-level class label $c$ are available. 
Thus, ISOD can be conducted by leveraging weakly-supervised object detection (WSOD) algorithms on the proposal representations and class labels obtained from \textsc{ImaginaryNet}. 
%

Following OICR\citep{tang2017multiple}, the object detection head consists of a \textbf{Multiple Instances Learning (MIL) branch} and a \textbf{Refinement branch} and the structure of them are shown in Fig~(\ref{fig:roi-head}). 
During the training phase, all object representations are first fed into the MIL branch. 
The MIL branch combines two parallel classification streams, each of which contains a softmax function. 
One softmax function is along the classes dimension to output the score matrix $\textbf{x}^m_c \in \mathbb{R}^{|\mathcal{P}| \times |\mathbb{C}|}$ , and the other is along the object dimension to output the score matrix $\textbf{x}^m_r \in \mathbb{R}^{|\mathcal{P}| \times |\mathbb{C}|}$. 
Subsequently, we compute the element-wise product of these two scores matrix and get the $\textbf{x}^m = \textbf{x}^m_c \odot \textbf{x}^m_r$.
$\textbf{x}^m$ indicates the final object-level classification scores. 
Then we get the $c$-th class image-level score $s^m_{c}$ by sum over all object: $s^m_c = \sum_{i=1}^{|\mathcal{P}|} x^{m}_{i,c}$ .
We convert the target label $c$ into one-hot form $\hat{\textbf{y}}$ and calculate the Binary Cross Entropy (BCE) loss:
\begin{equation}
    \mathcal{L}^{ign}_{mil} = -\sum\nolimits_{c=1}^{|\mathbb{C}|} \hat{y}_c \log{s^m_c} + (1-\hat{y}_c) \log{(1-s^m_c)}
    \label{mil loss}
\end{equation}
To further improve the performance of the detector, we integrate the self-training methodology and construct the Refinement branch, which has a similar structure as the RoI head of the R-CNN-like detector \citep{ren2015faster}.
Following \citep{tang2017multiple}, we keep the proposal with the highest classification score from $\textbf{x}^m$ as the pseudo-ground-truth (PGT), which are used to optimize the Refinement branch by a refinement loss $\mathcal{L}^{ign}_{ref}$. Finally, the learning objective of our object detection head of ISOD is: 
\begin{equation}
    \mathcal{L} = \mathcal{L}^{ign}_{mil} + \mathcal{L}^{ign}_{ref}
    \label{all loss}
\end{equation}
The whole training pipeline of ISOD is shown in Fig.~\ref{fig:roi-head}(a). Here we take the OICR as the example and it is feasible to employ other WSOD methods for the end of ISOD.

Furthermore, by gradually introducing real images and manual annotations, \textsc{ImaginaryNet} can collaborate with other supervision setting to boost detection performance. 
%
%
Here we present two examples by incorporating \textsc{ImaginaryNet} with weakly supervision and full supervision on real data.
During training, both real images and synthesized images are fed into the Representation Generator and object detection head in order and optimize these two modules via back-propagation. 
And we only need to slightly modify the training pipeline of object detection head. 
When real images and image-level annotations are available, \textit{i.e.}, $\mathcal{D}^\mathrm{R}_w \neq \emptyset$, as shown in Fig.~\ref{fig:roi-head}(b) we can  uniformly sample synthesized and real images and feed them into the WSOD training process.
%
%
When both real images and box-level annotations are available, \textit{i.e.}, $\mathcal{D}^\mathrm{R}_s \neq \emptyset$, we remove the MIL branch and treat the synthesis images as \textit{unlabeled image}.
As shown in Fig.~\ref{fig:roi-head}(c), we simply leverage the teacher-student framework widely adopted in SSOD to train the detection head. The teacher detection network generates pseudo-ground-truth (PGT) for unlabeled real images and imaginary images to train the student detection network. In addition, the student is also trained on real images with box-level ground-truth (GT) by a supervised-learning manner \citep{ren2015faster}.

%

\subsection{Training and Inference Pipeline}
%
During the training phase, \textsc{ImaginaryNet} samples a target class $c$ and generates the full description $\textbf{t}$ via a language model. Then we get an imaginary image by Eq.~(\ref{I}) as well as its feature map $\textbf{h}$ by $\textbf{h} = E^{\mathrm{I}}(\textbf{I})$. Consequently, a set of region proposals is generated by the Proposals Generator, and proposal representations are obtained via RoI Pooling. 
Finally, the proposal representations  are fed into different object detection heads according to the supervision settings. 
During the whole training process, only the parameters of the Representation Generator and Object Detection head need to be updated.
In the inference phase, the language model and synthesis model will be abandoned. 
Taking a real image as the input, the Representation Generator generates proposal representations, and the detection head decodes them to predict the detection bounding box.

\begin{table*}[t!]
    \caption{\textbf{Comparison of our \textsc{ImaginaryNet}(ISOD) with baseline methods on PASCAL VOC 2007.} We consider two baseline methods, a strong baseline based on unsupervised CLIP model, and an upper-bound baseline based on OICR trained on 5,000 real images with image-level annotations.}
    \centering
    \setlength{\tabcolsep}{2pt}
    \begin{sc}
    \resizebox{\textwidth}{!}{
    {
    \begin{tabular}{l c c c c c c c c c c c c c c c c c c c c c}
    \toprule
    Methods & Aero & Bike & Bird & Boat & Bottle & Bus & Car & Cat & Chair & Cow & Table & Dog & Horse & Motor & Person & Plant & Sheep & Sofa & Train & ~~TV~~ & mAP \\
    \midrule
    OICR (WSOD) &54.97&55.32&47.64&29.25&24.94&69.32&64.76&76.07&18.16&56.59&20.17&70.13&69.03&64.42&19.82&20.12&49.14&27.42&68.61&52.72&47.93\\
    \midrule
    CLIP &16.30&17.15&19.42&\textbf{8.80}&14.53&30.28&25.83&25.92&\textbf{20.34}&27.10&\textbf{21.85}&26.00&20.32&22.59&10.09&13.43&26.24&27.82&19.26&23.59&20.84\\
    ImaginaryNet & \textbf{47.18}&\textbf{43.13}&\textbf{35.51}&5.33&\textbf{19.23}&\textbf{32.58}&\textbf{41.23}&\textbf{66.72}&12.03&\textbf{43.98}&18.13&\textbf{43.98}&\textbf{36.71}&\textbf{51.68}&\textbf{19.04}&\textbf{14.97}&\textbf{34.12}&\textbf{42.74}&\textbf{23.04}&\textbf{33.36}&\textbf{33.23}\\
    \bottomrule
    \end{tabular}
    }}
    \end{sc}
    \vspace{-0.1in}
    \label{Main-ISOD}
    \end{table*}

%% file: Content/04_Experiment.tex
\section{Experiments}

\subsection{Implementation Details}

We use GPT-2 \citep{radford2019language} as the language model and DALLE-mini, which can better follow the language guidance, as the text-to-image synthesis model. 
We implement the image encoder with ResNet50 pretrained on ImageNet dataset. For Proposal Generator, we use Selective Search following W2N \citep{huang2022w2n} for reaching no real images and manual annotations. All training hyper-parameters follow OICR \citep{tang2017multiple} and W2N \citep{huang2022w2n} for a fair comparison. 
Unless otherwise specified, we generate 5,000 imaginary images during training, which has the similar amount of images in comparison to PASCAL VOC2007 trainval set.

\subsection{Experimental Setup}

We first compare \textsc{ImaginaryNet} with the ISOD model to verify whether it is feasible to learn object detectors without real images and manual annotations. 
To the best of our knowledge, we are among the first to investigate ISOD solely based on synthesized images. 
So we set up a strong baseline based on the CLIP model. 
For CLIP baseline, we use the Edge Boxes algorithm \footnote{We tried Selective Search but Edge Boxes is more effective, so we report the best results.} to extract potential proposals on VOC test images for CLIP to classify its class, based on its similarity score with pre-designed prompts. 
Before computing AP, we apply the Non-Maximum Suppression (NMS) operation to remove redundant proposals based on CLIP scores. See Appendix \ref{clip} for more details.

We also assess \textsc{ImaginaryNet} by collaborating with WSOD method W2N \citep{huang2022w2n}, \textit{i.e.}, \textsc{ImaginaryNet}(WSOD). For comparison, we consider WSDDN \citep{bilen2016weakly}, OICR \citep{tang2017multiple} and W2N \citep{huang2022w2n} on real images as the baseline detection networks. Compared with them, \textsc{ImaginaryNet} in ISOD, \textit{i.e.}, \textsc{ImaginaryNet}(ISOD), adopts similar training pipeline but does not require real images and manual annotations, whereas \textsc{ImaginaryNet}(WSOD) leverages both real and synthesized images.



\subsection{Overall Results}

\subsubsection{Imaginary-supervised Object Detection}
    
As shown in Tab. \ref{Main-ISOD}, \textsc{ImaginaryNet}(ISOD) achieves an mAP of 33.23. This verifies the feasibility of learning object detectors without real images and annotations. Moreover, \textsc{ImaginaryNet}(ISOD) outperforms the CLIP baseline model with a gain of 12.39 in mAP. This further supports that the competitiveness
of a pre-trained generative model against pre-trained contrastive alignment models (\textit{e.g.}, CLIP) for learning object detectors. 
%
%
For most classes, \textsc{ImaginaryNet}(ISOD) can obtain a significant performance gain, and \textsc{ImaginaryNet}(ISOD) has much fewer parameter amounts and computational cost in inference because the pre-trained modules can be abandoned after training. 
Especially, for some classes such as cat, the performance improved from 25.92 of CLIP to 66.72 of \textsc{ImaginaryNet}(ISOD).
%
%
The results clearly show the feasibility of \textsc{ImaginaryNet}(ISOD), \textit{i.e.}, learning object detectors without real images and manual annotations.  
%

We further compare \textsc{ImaginaryNet}(ISOD) with an upper-bound baseline, WSOD on real images with class-level annotations.
For a fair comparison, we only generate 5,000 imaginary samples during training.
Compared with OICR, which has the same backbone with Ours and trained on 5,000 real images and manual annotations in WSOD setting, we achieved about 70\% performance of it. 
Considering that synthesized images and image-level annotations are free for \textsc{ImaginaryNet}, our method has the potential to further improve detection accuracy by increasing training samples.

\paragraph{MSCOCO Results} We also conduct the experiments on MSCOCO dataset, and obtain similar observations as VOC2007. See Appendix \ref{coco} for more details.

\subsubsection{Incorporating with WSOD on Real Images and manual Annotations}

\begin{table*}[t!]
    \caption{\textbf{Comparison of \textsc{ImaginaryNet(WSOD)} on PASCAL VOC 2007.}
    All models use 5,000 real data from VOC2007 and \textsc{ImaginaryNet(WSOD)} leverages extra 5,000 imaginary samples.
    In comparison, \textsc{ImaginaryNet(WSOD)} achieves state-of-the-art performance.}
    \centering
    \setlength{\tabcolsep}{2pt}
    \begin{sc}
    \resizebox{\textwidth}{!}{
    {
    \begin{tabular}{l c c c c c c c c c c c c c c c c c c c c c c}
    \toprule
    Methods & Aero & Bike & Bird & Boat & Bottle & Bus & Car & Cat & Chair & Cow & Table & Dog & Horse & Motor & Person & Plant & Sheep & Sofa & Train & ~~TV~~ & mAP \\
    \midrule
    WSDDN & 39.41 & 50.09 & 31.48 & 16.30 & 12.61 & 64.45 & 42.82 & 42.63 & 10.06 & 35.72 & 24.92 & 38.24 & 34.41 & 55.60 & 9.39 & 14.72 & 30.22 & 40.68 & 54.70 & 46.94 & 34.77\\
    OICR &54.97&55.32&47.64&29.25&24.94&69.32&64.76&76.07&18.16&56.59&20.17&70.13&69.03&64.42&19.82&20.12&49.14&27.42&68.61&52.72&47.93\\
    W2N &74.00&73.82&59.40&28.34&43.43&\textbf{80.03}&72.61&81.23&14.00&\textbf{76.75}&25.98&58.64&63.76&\textbf{75.69}&10.86&29.56&\textbf{60.38}&63.76&79.56&67.29&56.95\\
    ImaginaryNet + WSDDN & 40.92&51.55&33.36&17.11&22.97&60.29&46.39&44.15&16.06&39.28&34.82&47.88&42.16&53.83&19.51&16.20&37.63&46.03&62.39&42.43&38.75 (+4.02)\\
    ImaginaryNet + OICR &59.65&65.84&49.54&30.01&32.52&68.67&64.30&76.57&25.19&55.5&40.20&65.96&63.86&65.81&27.78&26.31&50.97&56.24&70.23&65.85&53.05 (+5.12)\\
    ImaginaryNet + W2N & \textbf{74.66}&\textbf{81.36}&\textbf{64.72}&\textbf{37.40}&\textbf{49.97}&79.89&\textbf{74.79}&\textbf{84.23}&\textbf{34.31}&72.62&\textbf{49.59}&\textbf{79.27}&\textbf{76.99}&75.60&\textbf{40.52}&\textbf{34.80}&58.97&\textbf{64.82}&\textbf{81.60}&\textbf{72.81}&\textbf{64.45 (+7.50)}\\
    \bottomrule
    \end{tabular}
    }}
    \end{sc}
    \vskip -0.15in
    \label{Main-WSOD-SSOD-FSOD}
    \end{table*}
    
We further show that the synthesized images by \textsc{ImaginaryNet} can collaborate with other supervision settings on real images and manual annotations for boosting detection performance. %
Here we use WSOD as an example.  
As given in Tab. \ref{Main-WSOD-SSOD-FSOD}, \textsc{ImaginaryNet}(WSOD) has achieved state-of-the-art performance in WSOD. This shows the effectiveness of \textsc{ImaginaryNet} can not only learn object detectors without real images and annotations, but also is effective in improving the performance of backbones without increasing parameters and inference cost.
\textsc{ImaginaryNet}(WSOD) obtain an mAP of 64.45, which is higher than the state-of-the-art WSOD models.
%



\subsection{Comparison with Real Data}

\begin{table*}[t!]
    \caption{\textbf{Comparison of our method with real data on PASCAL VOC 2007.} Imaginary data can play a similar role in enhancing performance. 10K Imaginary data can improve the model by a similar margin compared with 10K real data. Moreover, introducing imaginary data when extra real data exists can further improve the performance.
    }
    \centering
    \begin{sc}
    \resizebox{\textwidth}{!}{
    {
    \begin{tabular}{l c c c c}
    \toprule
    Methods & Annotated Real Data & Un-annotated Real Data & Un-real Data & mAP \\
    \midrule
    FasterRCNN & 5K VOC2007&$\emptyset$&$\emptyset$&76.00\\
    Unbiased Teacher &5K VOC2007&10K VOC2012&$\emptyset$&81.08 (+5.08)\\
    \midrule
    ImaginaryNet(SSOD) &5K VOC2007&$\emptyset$& 5K Imaginary &80.30 (+4.30)\\
    ImaginaryNet(SSOD) &5K VOC2007&$\emptyset$& 10K Imaginary &80.59 (+4.49)\\
    ImaginaryNet(SSOD) &5K VOC2007&10K VOC2012& 10K Imaginary &82.00 (+6.00)\\
    \bottomrule
    \end{tabular}
    }}
    \end{sc}
    \vskip -0.15in
    \label{Comp-Real}
    \end{table*}

How effective is \textsc{ImaginaryNet} in comparison with real data? To answer this question, we conduct comparison experiments in SSOD setting. We use the same backbone, \textsc{ResNet50-FPN} as the supervised model \textsc{FasterRCNN} as well as the SSOD model \textsc{Unbiased Teacher}\footnote{Although the performance in its original paper is lower, we obtain 81.08 by re-training the model.}. For a fair comparison, we use the same backbone, training hyper-parameters, and real data by following the competing methods.

As shown in Tab. \ref{Comp-Real}, comparing with \textsc{Faster-RCNN} and \textsc{Unbiased Teacher}, \textsc{ImaginaryNet}(SSOD) has achieved comparable performance, about 99.3\% performance of \textsc{Unbiased Teacher}. This shows that generated data are almost as effective as real data but have much lower cost in data acquisition and higher controllability in class balance.
Compared to the baseline \textsc{Unbiased Teacher}, we can observe that generated data are orthogonal with real data, where we can improve the performance by increasing both real and synthesized data.

\subsection{Visualization}

To figure out why \textsc{ImaginaryNet} can learn object detectors without real images and manual annotations, we conduct several visualizations to explain the reason from different aspects.

\subsubsection{Extracted Features}

\begin{figure}
\begin{minipage}[t]{0.49\textwidth}
	\subfigure{
	\includegraphics[height=1.4in,width=2.6in]{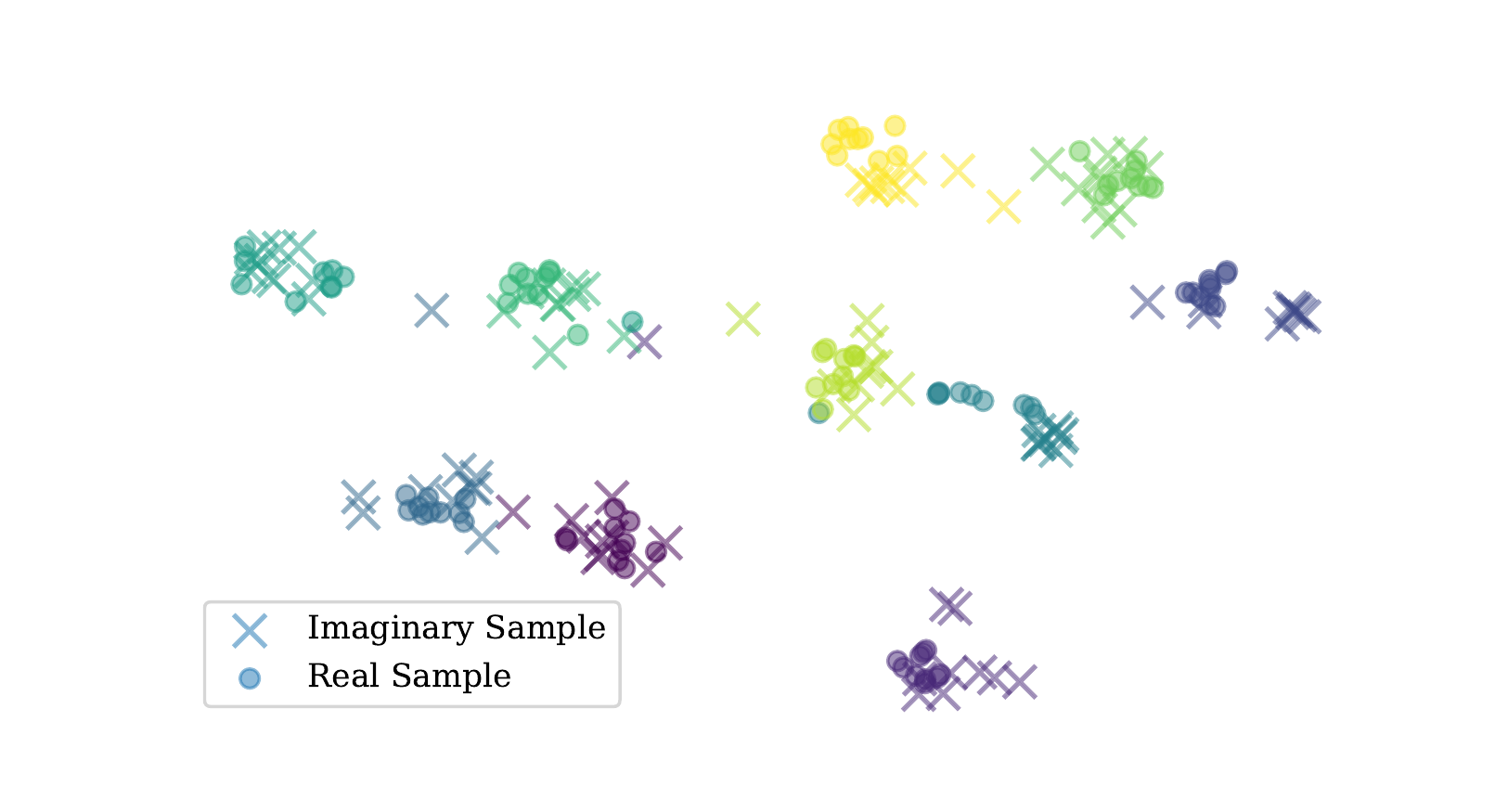}
	}
	\caption{\textbf{Visualization of extracted features.} We can observe that features can cluster based on the object type. This shows imaginary data contains similar knowledge like real data.}
 \label{cluster}
\end{minipage}
\hfill
\begin{minipage}[t]{0.49\textwidth}
	\subfigure{
	\includegraphics[width=2.6in]{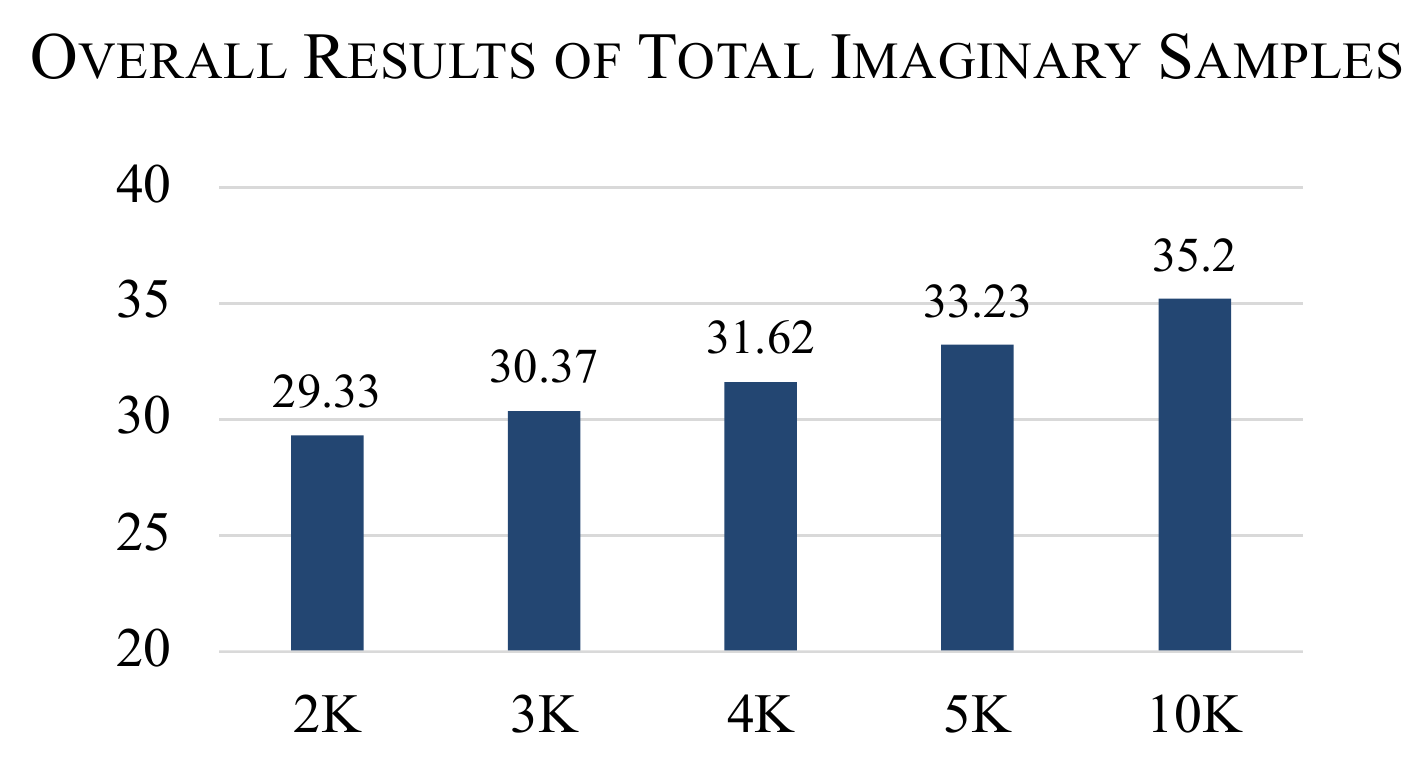}
	}
	\caption{\textbf{Overall results of different total imaginary samples.} The performance improves steadily with the growth of imaginary samples.}
\label{chart}
\end{minipage}
\vskip -0.15in
\end{figure}

To verify whether the features extracted from imaginary images contain similar information to real images, we extracted their features from the same Representation Generator. For real images, we extract 10 features per class as anchors and 10 features per class from imaginary images. We use T-SNE to project them to a 2D image.

In Fig. \ref{cluster}, we can observe that features of one class from imaginary images are coincident with the feature cluster of the same class from real images. 
This indicates that features from imaginary images contain similar information in comparison with real images, which explains why \textsc{ImaginaryNet} can learn object detectors even without real images and annotations.

\subsubsection{Imaginary Images}

\begin{figure*}
	\centering
	\includegraphics[width=14cm]{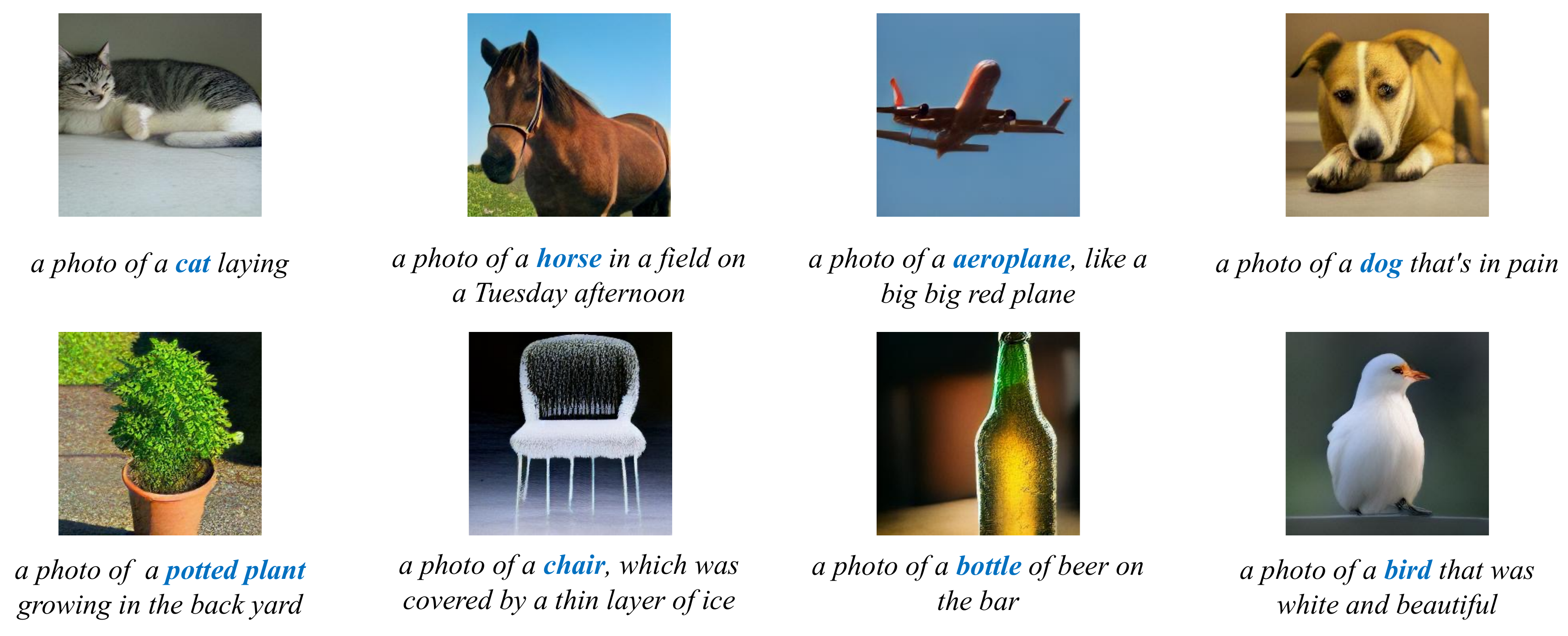}
	\caption{\textbf{Visualization of imaginary images.} Language model extends the prefix to a full description of the scene and the synthesis model generated images followed the description. }
	\label{fig:samples}
	\vskip -0.15in
\end{figure*}

We randomly sample imaginary images from the synthesis model and corresponding text descriptions from the language model to observe how they can play the role of real data. In Fig. \ref{fig:samples}, we can find that text descriptions are quite diverse. Meanwhile, high-quality images from the text-to-image synthesis model follow the description well and contain vivid objects of the corresponding class. Although all images generated are not in real datasets, \textsc{ImaginaryNet} successfully simulates the scene of the images from object detection datasets. This may explain why \textsc{ImaginaryNet} can even enhance object detectors under settings that contain real images and annotations.

\subsubsection{Case Studies}

\begin{figure*}
	\centering
	\includegraphics[width=14cm]{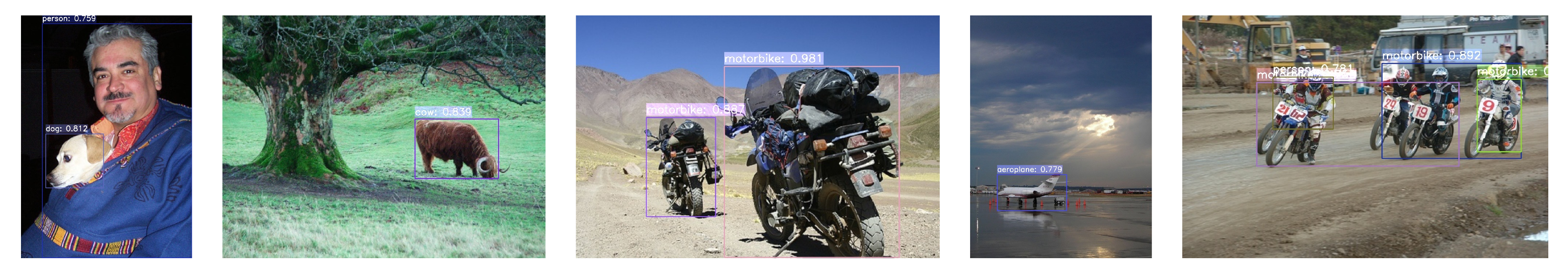}
	\caption{\textbf{Visualization of detection results on PASCAL VOC 2007 test split.} Results show that it is feasible to learn object detectors  via \textsc{ImaginaryNet}(ISOD).}
	\label{fig:case}
	\vskip -0.15in
\end{figure*}

To better reveal the quality of \textsc{ImaginaryNet}, we visualize detection results from VOC2007 test split in ISOD setting. In Fig. \ref{fig:case}, many objects are detected correctly. In some hard cases, \textit{e.g.}, the 5-th image which contains many drivers and motors overlapping together, can also be detected well with nearly-correct boxes and classes. This proves the effectiveness of \textsc{ImaginaryNet} in ISOD.

Although \textsc{ImaginaryNet} shows pretty effectiveness in many cases, we still find out some failure cases. For example, in Fig. \ref{fig:case}, some motorbikes are not detected well and persons are classified into the wrong class. This is because ISOD is a very strict setting, where the model cannot access any real images and annotations. Nonetheless, \textsc{ImaginaryNet} still shows that it is feasible to learn object detectors in this hard setting.

\section{Ablation Studies}

To further analyze the performance of \textsc{ImaginaryNet}, we conduct several ablation studies to investigate the proposed framework.

\subsection{Influence of Total Imaginary Samples}

\begin{table*}[t!]
    \caption{\textbf{Comparison of different imaginary samples in ISOD on PASCAL VOC 2007.} We set the different scales of samples and report their mAP. The mAP increases with the growth of samples.}
    \centering
    \setlength{\tabcolsep}{2pt}
    \begin{sc}
    \resizebox{\textwidth}{!}{
    {
    \begin{tabular}{l c c c c c c c c c c c c c c c c c c c c c}
    \toprule
    Samples & Aero & Bike & Bird & Boat & Bottle & Bus & Car & Cat & Chair & Cow & Table & Dog & Horse & Motor & Person & Plant & Sheep & Sofa & Train & ~~TV~~ & mAP \\
    \midrule
2K&39.35&43.68&33.36&4.51&13.84&29.39&38.71&59.48&10.69&35.75&21.06&37.43&28.97&43.64&17.20&13.55&33.10&35.71&17.51&29.73&29.33\\
    3K & 39.15&40.10&33.18&3.40&17.19&32.34&37.21&53.69&11.50&41.29&19.75&39.34&35.24&48.15&16.30&13.74&29.83&42.62&25.38&28.01&30.37\\
    4K &41.92&41.99&\textbf{35.64}&4.71&17.88&35.88&38.28&58.10&11.06&35.17&\textbf{24.19}&33.36&36.34&45.88&17.80&14.30&30.71&41.97&\textbf{26.45}&\textbf{40.82}&31.62\\
    5K &\textbf{47.18}&43.13&35.51&5.33&19.23&32.58&\textbf{41.23}&\textbf{66.72}&12.03&43.98&18.13&\textbf{43.98}&36.71&51.68&19.04&14.97&34.12&42.74&23.04&33.36&33.23\\
    10K &45.10&\textbf{51.31}&34.22&\textbf{5.34}&\textbf{27.65}&\textbf{36.62}&39.35&62.00&\textbf{13.41}&\textbf{45.31}&20.45&41.41&\textbf{51.13}&\textbf{52.25}&\textbf{21.32}&\textbf{21.48}&\textbf{35.20}&\textbf{44.48}&22.06&33.82&\textbf{35.20}\\
    \bottomrule
    \end{tabular}
    }}
    \end{sc}
    \vspace{-0.15in}
    \label{Ablation-Samples}
    \end{table*}

Can the model obtain higher performance if \textsc{ImaginaryNet} imagine more samples? To figure out this, we conduct experiments on ISOD with different scales of imaginary samples. In Tab. \ref{Ablation-Samples} and Fig. \ref{chart}, one can see that the overall performance, mAP, increases steadily to 35.20 finally. The result shows that more imaginary samples can make the detector stronger. Meanwhile, \textsc{ImaginaryNet} obtains mAP of 29.33 with 2K samples only. This shows the effectiveness of imaginary samples, which can train the detector with very limited samples. We also notice the fluctuation of several classes, which may be explained by the instability of class-level annotation learning. In general, more samples will improve the final performance. 

\subsection{Effectiveness of Language Model}

\begin{table*}[t!]
    \caption{\textbf{Ablation of language model in ISOD on PASCAL VOC 2007.} We can observe that the language model improve the results of the model significantly.}
    \centering
    \setlength{\tabcolsep}{2pt}
    \begin{sc}
    \resizebox{\textwidth}{!}{
    {
    \begin{tabular}{l c c c c c c c c c c c c c c c c c c c c c}
    \toprule
    Methods & Aero & Bike & Bird & Boat & Bottle & Bus & Car & Cat & Chair & Cow & Table & Dog & Horse & Motor & Person & Plant & Sheep & Sofa & Train & ~~TV~~ & mAP \\
    \midrule
    w/o LM & 31.32&\textbf{49.94}&32.13&3.24&\textbf{23.78}&\textbf{43.35}&34.66&30.95&\textbf{13.35}&38.59&15.77&38.51&\textbf{40.82}&42.34&15.84&\textbf{18.39}&\textbf{42.11}&38.86&18.99&\textbf{47.54}&31.02\\
    w/ LM &\textbf{47.18}&43.13&\textbf{35.51}&\textbf{5.33}&19.23&32.58&\textbf{41.23}&\textbf{66.72}&12.03&\textbf{43.98}&\textbf{18.13}&\textbf{43.98}&36.71&\textbf{51.68}&\textbf{19.04}&14.97&34.12&\textbf{42.74}&\textbf{23.04}&33.36&\textbf{33.23}\\
    \bottomrule
    \end{tabular}
    }}
    \end{sc}
    \vskip -0.15in
    \label{Ablation-LM}
    \end{table*}
    
Whether the various scene descriptions are useful for detector learning? To figure out this question, we re-train an \textsc{ImaginaryNet} without the language model. In Tab. \ref{Ablation-LM}, we can observe a significant drop in mAP. The performance of each class also drops in most cases. Such class, like cat, drops over 30 of AP50. The result indicates that language model is also important in generating diverse images. We also notice the unexpected drop in some classes and we believe this can be explained by the failure of the language model to generate a scene similar to the real dataset in some cases. We will leave this as future work to generate images that are more suitable for learning detectors. 

\subsection{Verification of Orthogonality with Data Augmentations}

\begin{table*}[t!]
    \caption{\textbf{Ablation of data augmentation strategies training on PASCAL VOC 2007.} We can observe that all augmentation strategies are effective on imaginary samples.}
    \centering
    \setlength{\tabcolsep}{2pt}
    \begin{sc}
    \resizebox{\textwidth}{!}{
    {
    \begin{tabular}{c c c c c c c c c c c c c c c c c c c c c c c}
    \toprule
    Train Aug. & Test Aug. & Aero & Bike & Bird & Boat & Bottle & Bus & Car & Cat & Chair & Cow & Table & Dog & Horse & Motor & Person & Plant & Sheep & Sofa & Train & ~~TV~~ & mAP \\
    \midrule
     &  & 86.05&86.75&81.92&66.66&71.11&86.72&88.13&87.68&61.24&84.58&74.15&87.37&86.73&86.95&85.65&56.45&81.45&76.94&85.03&76.89&79.92\\
     $\surd$ &  & 86.84&87.55&82.99&70.01&73.50&86.85&88.45&87.56&62.07&83.92&73.88&87.26&87.45&87.62&85.76&57.09&81.14&78.93&84.51&78.42&80.59\\
     $\surd$ & $\surd$ & 87.83&88.55&84.83&76.29&76.08&87.68&89.19&88.28&66.40&87.34&75.66&88.52&88.77&88.62&86.58&62.33&82.23&79.86&87.71&81.22&82.70 \\
    \bottomrule
    \end{tabular}
    }}
    \end{sc}
    \vskip -0.15in
    \label{table:aug}
    \end{table*}

Whether the data augmentation strategies still work on imaginary images? To answer the question, we conduct several ablation studies with different data augmentation strategies, training augmentation, and testing augmentation.
For most WSOD method, which are also employed into our ISOD setting, no training augmentation is needed because it is relatively hard to learn with class-labeled images. Consequently, we conduct this experiment in SSOD setting.
As Tab. \ref{table:aug} shows, we can observe that \textsc{ImaginaryNet}(SSOD) obtains better performance after activating training augmentation and testing augmentation.
These results give strong evidence that our framework is orthogonal with data augmentations.
Moreover, orthogonality with data augmentation strategies may imply that imaginary samples have a similar property to real data.

%% file: Content/05_Conclusion.tex
\section{Broader Impact}

\textsc{ImaginaryNet} is an effective framework to learn detectors without real images and annotations for further improving the performance with real data. It can provide new insight for handling complex vision tasks under low-resource regime.
For example, it is possible to learn detection or segmentation models without any real data. 
However, the performance of \textsc{ImaginaryNet}(ISOD) is still inferior to WSOD and FSOD. 
Thus, more studies are required to generate diverse and photo-realistic images and to better extract supervision information from the language descriptions and synthesized images.


\section{Conclusion}

In this paper, we proposed a novel learning paradigm of object detection, Imaginary-Supervised Object Detection (ISOD), where no real images and manual annotations can be used for training object detectors.
In particular, we presented an \textsc{ImaginaryNet} framework to
generate synthesized images as well as supervision information.
Experiments shows that \textsc{ImaginaryNet}(ISOD) can obtain about 70\% of performance in comparison with the WSOD counterpart trained using real data and image-level annotations.
Moreover, \textsc{ImaginaryNet} can collaborate with other supervision settings on real data to further boost detection performance.

%% file: Content/0A_Appendix.tex
\section{Appendix}


\subsection{CLIP Baseline}
\label{clip} 
We use ViT-B/16 as the image encoder for CLIP and use  Edge Boxes to extract 1000 proposals for each test image. Each proposal will be transformed to $384 \times 384$ to fit the CLIP input size. Then each proposal will be calculated similarity with the prompt which consists of the prefix "a photo of" and the class name. We select the highest CLIP similarity as the class of the proposal.  Before computing AP, we apply the Non-Maximum Suppression (NMS) operation to remove redundant proposals according to their CLIP scores.

\subsection{Imaginary-supervised Object Detection Results on MSCOCO}
\label{coco}
\begin{table*}[h]
    \caption{\textbf{Comparison of our method in ISOD on MSCOCO2014.} We obtain a much higher mAP compared with the CLIP baseline. This shows the feasibility of training the detector on a harder dataset of 80 classes without any real images and annotations.}
    \centering
    \begin{sc}
    {
    \begin{tabular}{l c}
    \toprule
    Methods & mAP \\
    \midrule
    CLIP & 6.3 \\
    ImaginaryNet & \textbf{11.4} \\
    \bottomrule
    \end{tabular}
    }
    \end{sc}
    \label{ISOD-COCO}
    \end{table*}

We also verified our framework on MSCOCO2014, which is also a common dataset in object detection. As the results in Tab. \ref{ISOD-COCO}, we outperform the CLIP baseline with nearly two times of mAP. This proves that our framework is also effective on the harder dataset, which contains 80 classes.

\subsection{Comparison of Different Visual Synthesis Model}

\begin{table*}[h]
    \caption{\textbf{Comparison of different visual synthesis models in ISOD on PASCAL VOC 2007.} We can observe that \textsc{Dalle-mini} can provide higher performance for \textsc{ImaginaryNet}.}
    \centering
    \setlength{\tabcolsep}{2pt}
    \begin{sc}
    \resizebox{\textwidth}{!}{
    {
    \begin{tabular}{l c c c c c c c c c c c c c c c c c c c c c}
    \toprule
    Methods & Aero & Bike & Bird & Boat & Bottle & Bus & Car & Cat & Chair & Cow & Table & Dog & Horse & Motor & Person & Plant & Sheep & Sofa & Train & ~~TV~~ & mAP \\
    \midrule
    Stable Diffusion &35.71&24.26&13.67&14.96&1.17&\textbf{34.72}&31.76&41.43&9.76&20.66&25.39&36.04&27.37&35.90&10.46&3.78&17.50&36.68&30.89&13.53&23.28 \\
    Dalle-mini & \textbf{47.18}&\textbf{43.13}&\textbf{35.51}&\textbf{5.33}&\textbf{19.23}&32.58&\textbf{41.23}&\textbf{66.72}&\textbf{12.03}&\textbf{43.98}&\textbf{18.13}&\textbf{43.98}&\textbf{36.71}&\textbf{51.68}&\textbf{19.04}&\textbf{14.97}&\textbf{34.12}&\textbf{42.74}&\textbf{23.04}&\textbf{33.3}6&\textbf{33.23}\\
    \bottomrule
    \end{tabular}
    }}
    \end{sc}
    \label{Comp-Gen}
    \end{table*}

We noticed that the generated images are different from different generative models. In this paper, we use two effective open-source visual synthesis models, \textsc{Stable Diffusion} and \textsc{Dalle-mini}. We conduct a comparison on ISOD to verify the influence of synthesis models. In Tab. \ref{Comp-Gen}, we can find that \textsc{Dalle-mini} can provide higher performance for \textsc{ImaginaryNet}. We explained this as \textsc{Dalle-mini}'s results contain fewer objects, which are easier for the model to learn in the training.

\subsection{Data Ensemble}

To obtain a better performance in WSOD, the model trained with selected classes of Imaginary samples instead of all Imaginary samples. We first trained the model with all imaginary samples and evaluate the performance of each class in val split. We selected classes that the performance is higher than baseline model. Then we re-trained the model with imaginary samples of selected classes to obtain the best performance. As the results in Tab. \ref{ens}, we can observe that the \textsc{ImaginaryNet} obtain a better performance even without data ensemble. Combining with data ensemble, we obtain the best performance. We explained this as the gap among Imaginary samples and real samples in some classes, such as boat or train.

\begin{table*}[h]
    \caption{\textbf{Comparison of data used in WSOD on PASCAL VOC 2007.} We can observe that data ensemble improve the performance of \textsc{ImaginaryNet} significantly.}
    \centering
    \setlength{\tabcolsep}{2pt}
    \begin{sc}
    \resizebox{\textwidth}{!}{
    {
    \begin{tabular}{l c c c c c c c c c c c c c c c c c c c c c c}
    \toprule
    Methods & Data Ensemble & Aero & Bike & Bird & Boat & Bottle & Bus & Car & Cat & Chair & Cow & Table & Dog & Horse & Motor & Person & Plant & Sheep & Sofa & Train & ~~TV~~ & mAP \\
    \midrule
    OCIR & & 54.97 & 55.32 & 47.64 & 29.25 & 24.94 & \textbf{69.32} & \textbf{64.76} & 76.07 & 18.16 & 56.59 & 20.17 & \textbf{70.13} & 69.03 & 64.42 & 19.82 & 20.12 & 49.14 & 27.42 & 68.61 & 52.72 & 47.93 \\
    ImaginaryNet + OCIR & &57.45 & 60.68 & 46.65 & 9.93 & \textbf{33.11} & 66.27 & 63.20 & 72.68 & 19.75 & \textbf{57.19} & 26.14 & 66.84 & 65.75 & 62.99 & \textbf{35.36} & \textbf{27.21} & 47.96 & 53.49 & 48.43 & \textbf{66.43} & 49.38 (+1.45) \\
    ImaginaryNet + OCIR & $\checkmark$ &\textbf{59.65} & \textbf{65.84} & \textbf{49.54} & \textbf{30.01} & 32.52 & 68.67 & 64.30 & \textbf{76.57} & \textbf{25.19} & 55.50 & \textbf{40.20} & 65.96 & \textbf{63.86} & \textbf{65.81} & 27.78 & 26.31 & \textbf{50.97} & \textbf{56.24} & \textbf{70.23} & 65.85 & \textbf{53.05 (+5.12)} \\
    \bottomrule
    \end{tabular}
    }}
    \end{sc}
    \label{ens}
    \end{table*}